\documentclass{article}

\usepackage{microtype}
\usepackage{graphicx}
\usepackage{subcaption}
\usepackage{booktabs} %

\usepackage{hyperref}

\usepackage[preprint]{icml2026}

\usepackage{tikz}
\usetikzlibrary{arrows.meta}
\usepackage{pgfplots}

\usepackage[utf8]{inputenc} %
\usepackage[T1]{fontenc}    %
\usepackage{hyperref}       %
\usepackage{url}            %
\usepackage{booktabs}       %
\usepackage{amsfonts}       %
\usepackage{nicefrac}       %
\usepackage{microtype}      %
\usepackage{xcolor}         %

\usepackage{graphicx}
\usepackage{xspace}
\usepackage{subcaption}
\usepackage[normalem]{ulem}

\usepackage{algorithmic}
\usepackage{algorithm}
\usepackage{amsmath}
\usepackage{amssymb}
\usepackage{mathtools}
\usepackage{amsthm}

\usepackage{multirow}
\usepackage{wrapfig} 
\usepackage{colortbl}

\usepackage{amsmath,amsfonts,bm}

\def\eqref#1{equation~\ref{#1}}

\def\1{\bm{1}}

\def\rmI{{\mathbf{I}}}

\def\rmS{{\mathbf{S}}}

\def\vk{{\bm{k}}}

\def\vo{{\bm{o}}}

\def\vq{{\bm{q}}}

\def\vv{{\bm{v}}}

\DeclareMathAlphabet{\mathsfit}{\encodingdefault}{\sfdefault}{m}{sl}
\SetMathAlphabet{\mathsfit}{bold}{\encodingdefault}{\sfdefault}{bx}{n}

\def\gI{{\mathcal{I}}}

\def\gZ{{\mathcal{Z}}}

\newcommand{\E}{\mathbb{E}}

\newcommand{\R}{\mathbb{R}}

\newcommand{\softmax}{\mathrm{softmax}}

\providecommand{\select}{\text{Select}}
\providecommand{\bsa}{\text{SparseAttn}}
\providecommand{\cov}{\text{Cov}}
\providecommand{\blockscore}{\text{BlockScore}}

\newcommand{\sysname}{\textsc{SPLA}\xspace}
\newcommand{\sysnameablate}{\textsc{SPA}\xspace}

\providecommand{\rms}{\text{RMS}}

\usepackage[capitalize,noabbrev]{cleveref}

\theoremstyle{plain}

\theoremstyle{definition}

\theoremstyle{remark}

\usepackage[textsize=tiny]{todonotes}

\icmltitlerunning{\sysname: Block Sparse Plus Linear Attention for Long Context Modeling}

\begin{document}

\twocolumn[
  \icmltitle{\sysname: Block Sparse Plus Linear Attention for \\
    Long Context Modeling}

  \icmlsetsymbol{equal}{*}

  \begin{icmlauthorlist}
    \icmlauthor{Bailin Wang}{apple}
    \icmlauthor{Dan Friedman }{apple}
    \icmlauthor{Tao Lei}{apple}
    \icmlauthor{Chong Wang}{apple}
  \end{icmlauthorlist}

  \icmlaffiliation{apple}{Apple, California, USA}

  \icmlcorrespondingauthor{Bailin Wang}{bailin.wang28@gmail.com}

  \icmlkeywords{Machine Learning, ICML}

  \vskip 0.3in
]

\printAffiliationsAndNotice{Work done while the first and last authors were at Apple.}  %

\begin{abstract}
Block-wise sparse attention offers significant efficiency gains for long-context modeling, yet existing methods often suffer from low selection fidelity and cumulative contextual loss by completely discarding unselected blocks. To address these limitations, we introduce Sparse Plus Linear Attention (SPLA), a framework that utilizes a selection metric derived from second-order Taylor expansions to accurately identify relevant blocks for exact attention. Instead of discarding the remaining ``long tail,'' SPLA compresses unselected blocks into a compact recurrent state via a residual linear attention (RLA) module. Crucially, to avoid IO overhead, we derive an optimized subtraction-based formulation for RLA—calculating the residual as the difference between global and selected linear attention—ensuring that unselected blocks are never explicitly accessed during inference. Our experiments demonstrate that SPLA closes the performance gap in continual pretraining, surpassing dense attention models on long-context benchmarks like RULER while maintaining competitive general knowledge and reasoning capabilities.
\end{abstract}

\section{Introduction}
The capabilities of Large Language Models (LLMs) have scaled rapidly,
with a particular emphasis on extending context windows to millions of tokens.
As sequence lengths increase, however, the decoding phase becomes a critical bottleneck.
Because decoding is memory-bound,
serving long-context models with full attention requires loading increasingly large Key-Value (KV) caches,
often necessitating the usage of more GPUs/TPUs to accommodate the memory footprint.

To alleviate this, blockwise sparse attention mechanisms (e.g., NSA~\cite{nsa}, InfLLM~\cite{inf-v2}) have emerged as a promising optimization.
By selectively loading only the most relevant blocks of the cache,
these methods significantly increase inference throughput and reduce latency.
However, current sparse approaches face two fundamental limitations. First, their block selection methods—which are based on highly compressed block representations—suffer from \textit{low selection fidelity}, often failing to accurately identify the most relevant blocks. Second, by strictly truncating unselected blocks, existing methods suffer from \textit{cumulative contextual loss}. As context grows, the probability mass in the ``long tail'' becomes non-negligible, introducing a divergence between sparse and dense attention that severely degrades generation quality.

In this paper, we introduce \sysname,
a principled framework designed to close the performance gap between sparse and dense attention.
First, \sysname leverages second-order Taylor expansions
to derive a selection metric directly from a principled approximation of the original attention objective.
Furthermore, instead of simply truncating the context,
we introduce a \textit{Sparse Plus Linear Attention} paradigm.
We partition the context into "exact" and "approximate" sets:
relevant blocks are retrieved for exact attention,
while the ``long tail" is compressed into a recurrent state via \textit{residual linear attention} (RLA). Crucially, we implement RLA via a subtraction-based formulation that requires \textbf{no explicit memory access} for unselected blocks, maintaining the IO profile of sparse decoding.

In summary, the contributions of this work are as follows:
\begin{enumerate}
    \item We derive a principled block selection strategy based on second-order Taylor expansions. By utilizing block-level mean and covariance statistics, we bridge the mathematical gap between token-level attention and block-level retrieval, enabling more accurate selection.
    \item We propose \sysname, a novel framework that strictly partitions the context into exact and approximate components. By processing unselected blocks via residual linear attention, \sysname preserves the full global context---closing the performance gap with dense attention on sequence lengths up to 256k while retaining the efficiency of sparse decoding.
\end{enumerate}

\sysname is particularly well-suited for adapting pretrained dense attention models.
Empirically, we demonstrate that dense models can be converted to \sysname architectures
with a negligible number of additional parameters.
We believe this represents a promising direction for efficient long-context modeling:
standard attention architectures can be maintained during the expensive pretraining phase,
and then seamlessly adapted into efficient sparse variants during the continual pretraining stage.
\section{Background and Motivation}
\label{sec:background}

We use the decoding workload to describe the framework and motivation of the sparse attention mechanism, though these concepts generalize to other primary workloads such as training and prefilling.
In the decoding phase, the attention mechanism computes the output $\vo$ as:
\begin{equation}
    \vo = \sum_{i=1}^N \frac{\exp{(\vq \vk_i^\intercal)}}{\gZ(\vq, \vk_{1:N})}  \vv_i
\end{equation}
where $\vq, \vk_i, \vv_i \in \R^d$ denote the query, key, and value vectors, respectively, and $\gZ$ denotes the normalization term.
Conventionally, $\vk_{1:N}$ and $\vv_{1:N}$ are referred to as the KV cache, which must be retained in memory to be attended to by all future query vectors.
For simplicity, we assume the current query token is at position $N$, requiring consideration of all preceding tokens in the cache.

The decoding workload is intrinsically memory-bound.
The majority of runtime is dominated by loading the KV cache from memory, while the actual computation (vector dot-products) is relatively cheap, resulting in low arithmetic intensity.
Consequently, when the sequence length $N$ becomes large (e.g., 1 million tokens), decoding efficiency degrades significantly, particularly in terms of latency.
The central idea of sparse attention is to alleviate this memory bottleneck by selectively loading only a subset of the KV cache for computation.
In this section, we describe the selection mechanism and the subsequent sparse attention procedure.

\paragraph{Notation.} We denote the page (or block) size as $B$, the group size for Grouped Query Attention (GQA) as $G$, the number of selected pages as $k$, the head dimension as $d$, and the sequence length as $N$. $\gZ$ denotes the normalization function.

\subsection{Page Attention and Selection}

In practice, attention modules are typically served using logical pages (i.e., PagedAttention~\cite{pageattn}) to maximize memory utilization and continuity.
Given this memory layout, it is natural to select KV caches at the granularity of the page size $B$.
Specifically, the most relevant pages are identified to perform attention:
\begin{equation}
    \vo = \sum_{i \in \gI} \frac{\exp{(\vq \vk_{[i]}^\intercal)}}{\gZ(\vq, \vk_{\gI})}  \vv_{[i]} \in \R^{d}
    \label{eq:dense},
\end{equation}
where $\vk_{[i]} := \vk_{(i-1)B+1 : iB}$ and $\vv_{[i]} := \vv_{(i-1)B+1 : iB} \in \R^{B \times d}$ represent the key and value blocks, and $\gI \subset [N/B]$ denotes the set of indices of the relevant blocks.
The index set $\gI$ is computed via a selection operator applied to block summaries:
\begin{equation}
    \gI = \select \big (\{ \blockscore(\vq, \vk_{[i]}) \}_{i=1}^{N/B} \big ) 
    \label{eq:sparse-attn}
\end{equation}
where 
\begin{equation*}
\blockscore(\vq, \vk_{[i]}) := \frac{\exp{(\vq \tilde \vk_{[i]}^\intercal)}}{\gZ(\vq, \tilde \vk)}  \quad \in \R \quad .
\end{equation*}
Here, $\tilde \vk_{[i]}$ is a mapping that obtains a compact, block-level representation from the full block $\vk_{[i]}$.
The selector is designed to be input-dependent, meaning $\gI$ is a function of $\vq$, unlike input-agnostic patterns such as sliding windows~\cite{streaming}.

Existing works explore various designs for these mappings and selection functions.
Regarding the mapping, Quest~\cite{quest} utilizes non-parametric pooling operations (specifically min and max values), while Inf-LLM v2~\cite{inf-v2} employs mean pooling.
In contrast, NSA~\cite{nsa} utilizes a parametric MLP layer to learn block representations.
Regarding the selection function, TopK~\cite{nsa,inf-v2} is the most common choice due to its hardware efficiency.
Alternatively, TopP~\cite{twillight} employs nucleus sampling to offer flexibility when the attention distribution is long-tailed, while other approaches~\cite{vattention} propose hybrids of top-$k$ and random sampling.
Some methods, such as \cite{magicpig}, utilize Locality Sensitive Hashing (LSH), though this is often suboptimal as it is data-agnostic.

\subsection{Grouped Sparsity}

To generalize from single-head to multi-head settings, it is efficient to enforce a shared sparsity pattern across multiple query heads within the same group, consistent with Grouped Query Attention (GQA)~\cite{gqa}.
In this setting, the output $\vo^\text{\scriptsize{sparse}}$ is defined as:
\begin{align}
\vo^\text{\scriptsize{sparse}} &= \bsa(\vq, \vk_{1:N}, \vv_{1:N})  \\
& = \sum_{i \in \gI} \frac{\exp{(\vq \vk_{[i]}^\intercal)}}{\gZ(\vq,  \vk_{1:N}, \gI)}  \vv_{[i]} \quad \in \R^{G \times d},
    \label{eq:gqa-sparse}
\end{align}
where we overload the notation $\vq \in \R^{G \times d}$ to represent the concatenation of the $G$ query vectors in the group.
Consequently, the term $\vq \vk_{[i]}^\intercal$ yields a $G \times B$ score matrix, and the normalization function $\gZ$ operates row-wise.
From a computational perspective, such grouped sparsity improves hardware utilization by enabling parallelization along the head dimension.

To identify the most relevant pages for the entire group, we aggregate (sum) the relevance scores over all $G$ query heads:
\begin{equation}
  \blockscore(\vq, \vk_{[i]}) := \sum_{m=1}^G \frac{\exp{(\vq^m \tilde \vk_{[i]}^\intercal)}}{\gZ(\vq^m, \tilde \vk_{1:N/B})} \quad \in \R,
    \label{eq:block_score}
\end{equation}
where $\select$ provides the selected indices based on the score function $\blockscore$ with respect to the block index $i$.
Intuitively, during decoding, the majority of blocks are skipped; only their compressed representations $\tilde \vk_{[i]}$ are accessed for selection.
In the remainder of this paper, we use the term \textit{blocks} to refer to pages, as the presented techniques generalize to scenarios where KV caches are not arranged as pages.

\subsection{Open Challenges}

As a key motivation for this work, we identify two fundamental limitations within the current sparse attention framework:

\paragraph{1. Block Selection Fidelity.}
Current selection metrics are often heuristic approximations that lack a clear mathematical connection to the original token-level objective.
For instance, non-parametric pooling (min/max) is efficient but prone to selection errors, while parametric predictors (MLPs) require expensive training and struggle to generalize.
This lack of a \textit{principled} selection metric leads to high recall errors, where the model fails to retrieve semantically critical blocks.

\paragraph{2. Long Tail Divergence.}
As context length increases, the number of relevant blocks naturally grows.
Enforcing a fixed sparsity budget (constant $k$) necessitates discarding an increasing amount of probability mass located in the "long tail" of the distribution.
Simply truncating these blocks ($\vo_{i \notin \gI} = 0$) causes the sparse attention output to diverge progressively from the dense baseline, leading to significant degradation in model quality.

\section{\sysname: Block Sparse Attention with Residual Linear Attention}

To address these limitations, we present \sysname by formulating it as a decomposition of the token interactions into two complementary components:
\begin{align*}
    \sysname & = \underbrace{\text{Sparse Attention (Exact)}}_{\text{High-Relevance Peaks}} + \\ & \hspace{4mm} \underbrace{\text{Linear Attention (Approximate)}}_{\text{Low-Relevance Tail}}.
\end{align*}
This framework allows us to (1) use a Taylor expansion to identify the "peaks" (selected blocks) without heuristics, and (2) use Residual Linear Attention to efficiently compress and retain the "tail" (unselected blocks) instead of discarding it.

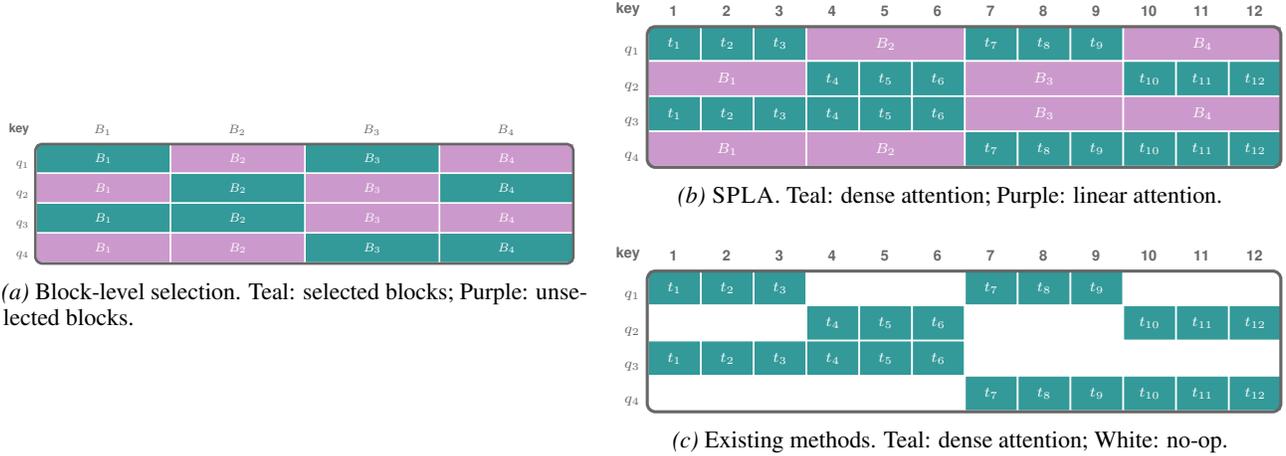
\begin{figure*}[t!]
    \centering

    \begin{subfigure}[c]{0.45\linewidth}
        \centering
        \resizebox{\linewidth}{!}{%
            \begin{tikzpicture}[
    font=\sffamily\bfseries\scriptsize,
    x=0.9cm, y=0.6cm,
    selected_block/.style={
        draw=white, 
        line width=0.8pt, 
        fill=teal!80!white, 
        text=white, 
        minimum width=2.7cm, 
        minimum height=0.6cm, 
        anchor=south west
    },
    unselected_block/.style={
        draw=white, 
        line width=0.8pt, 
        fill=violet!40, 
        text=white, 
        minimum width=2.7cm, 
        minimum height=0.6cm, 
        anchor=south west
    },
    label/.style={
        text=gray!80!black,
        font=\sffamily\bfseries\scriptsize,
        anchor=east
    },
    collabel/.style={
        text=gray!80!black,
        font=\sffamily\bfseries\scriptsize,
        anchor=south
    }
]

    \node[collabel, anchor=south east] at (0, 4.1) {key};

    \foreach \k in {1,2,3,4} {
        \node[collabel] at (3*\k-1.5, 4.1) {$B_{\k}$}; 
    }

    \node[label] at (0, 3.3) {$q_1$};
    \node[selected_block] at (0, 3) {$B_1$};
    \node[unselected_block] at (3, 3) {$B_2$};
    \node[selected_block] at (6, 3) {$B_3$};
    \node[unselected_block] at (9, 3) {$B_4$};

    \node[label] at (0, 2.3) {$q_2$};
    \node[unselected_block] at (0, 2) {$B_1$};
    \node[selected_block] at (3, 2) {$B_2$};
    \node[unselected_block] at (6, 2) {$B_3$};
    \node[selected_block] at (9, 2) {$B_4$};

    \node[label] at (0, 1.3) {$q_3$};
    \node[selected_block] at (0, 1) {$B_1$};
    \node[selected_block] at (3, 1) {$B_2$};
    \node[unselected_block] at (6, 1) {$B_3$};
    \node[unselected_block] at (9, 1) {$B_4$};

    \node[label] at (0, 0.3) {$q_4$};
    \node[unselected_block] at (0, 0) {$B_1$};
    \node[unselected_block] at (3, 0) {$B_2$};
    \node[selected_block] at (6, 0) {$B_3$};
    \node[selected_block] at (9, 0) {$B_4$};

    \draw[gray!80!black, line width=1.5pt, rounded corners=4pt] 
        (0,0) rectangle (12,4);

\end{tikzpicture}
        }
        \caption{Block-level selection. Teal: selected blocks; Purple: unselected blocks.}
        \label{fig:nsa_block}
    \end{subfigure}
    \hfill
    \begin{minipage}[c]{0.53\linewidth}

        \begin{subfigure}[b]{\linewidth}
            \centering
            \resizebox{\linewidth}{!}{%
                \begin{tikzpicture}[
    font=\sffamily\bfseries\scriptsize,
    x=0.9cm, y=0.6cm,
    token/.style={
        draw=white, 
        line width=0.8pt, 
        fill=teal!80!white,
        text=white, 
        minimum width=0.9cm, 
        minimum height=0.6cm, 
        anchor=south west
    },
    block/.style={
        draw=white, 
        line width=0.8pt, 
        fill=violet!40,
        text=white, 
        minimum width=2.7cm, 
        minimum height=0.6cm, 
        anchor=south west
    },
    label/.style={
        text=gray!80!black,
        font=\sffamily\bfseries\scriptsize,
        anchor=east
    },
    collabel/.style={
        text=gray!80!black,
        font=\sffamily\bfseries\scriptsize,
        anchor=south
    }
]

    \node[collabel, anchor=south east] at (0, 4.1) {key};

    \foreach \col in {1,...,12} {
        \node[collabel] at (\col-0.5, 4.1) {\col};
    }

    \node[label] at (0, 3.3) {$q_1$};
    \foreach \i in {1,2,3} { \node[token] at (\i-1, 3) {$t_{\i}$}; }
    \node[block] at (3, 3) {$B_2$};
    \foreach \i in {7,8,9} { \node[token] at (\i-1, 3) {$t_{\i}$}; }
    \node[block] at (9, 3) {$B_4$};

    \node[label] at (0, 2.3) {$q_2$};
    \node[block] at (0, 2) {$B_1$};
    \foreach \i in {4,5,6} { \node[token] at (\i-1, 2) {$t_{\i}$}; }
    \node[block] at (6, 2) {$B_3$};
    \foreach \i in {10,11,12} { \node[token] at (\i-1, 2) {$t_{\i}$}; }

    \node[label] at (0, 1.3) {$q_3$};
    \foreach \i in {1,2,3} { \node[token] at (\i-1, 1) {$t_{\i}$}; }
    \foreach \i in {4,5,6} { \node[token] at (\i-1, 1) {$t_{\i}$}; }
    \node[block] at (6, 1) {$B_3$};
    \node[block] at (9, 1) {$B_4$};

    \node[label] at (0, 0.3) {$q_4$};
    \node[block] at (0, 0) {$B_1$};
    \node[block] at (3, 0) {$B_2$};
    \foreach \i in {7,8,9} { \node[token] at (\i-1, 0) {$t_{\i}$}; }
    \foreach \i in {10,11,12} { \node[token] at (\i-1, 0) {$t_{\i}$}; }

    \draw[gray!80!black, line width=1.5pt, rounded corners=4pt] 
        (0,0) rectangle (12,4);

\end{tikzpicture}
            }
            \caption{\sysname. Teal: dense attention; Purple: linear attention.}
            \label{fig:hybrid}
        \end{subfigure}

        \vspace{1em}

        \begin{subfigure}[b]{\linewidth}
            \centering
            \resizebox{\linewidth}{!}{%
                \begin{tikzpicture}[
    font=\sffamily\bfseries\scriptsize,
    x=0.9cm, y=0.6cm,
    token/.style={
        draw=white, 
        line width=0.8pt, 
        fill=teal!80!white, 
        text=white, 
        minimum width=0.9cm, 
        minimum height=0.6cm, 
        anchor=south west
    },
    empty_block/.style={
        draw=white, %
        line width=0.8pt, 
        fill=white, 
        text=gray!50, %
        minimum width=2.7cm, 
        minimum height=0.6cm, 
        anchor=south west
    },
    label/.style={
        text=gray!80!black,
        font=\sffamily\bfseries\scriptsize,
        anchor=east
    },
    collabel/.style={
        text=gray!80!black,
        font=\sffamily\bfseries\scriptsize,
        anchor=south
    }
]

    \node[collabel, anchor=south east] at (0, 4.1) {key};
    \foreach \col in {1,...,12} {
        \node[collabel] at (\col-0.5, 4.1) {\col};
    }

    \node[label] at (0, 3.3) {$q_1$};
    \foreach \i in {1,2,3} { \node[token] at (\i-1, 3) {$t_{\i}$}; } %
    \node[empty_block] at (3, 3) {}; %
    \foreach \i in {7,8,9} { \node[token] at (\i-1, 3) {$t_{\i}$}; } %
    \node[empty_block] at (9, 3) {}; %

    \node[label] at (0, 2.3) {$q_2$};
    \node[empty_block] at (0, 2) {}; %
    \foreach \i in {4,5,6} { \node[token] at (\i-1, 2) {$t_{\i}$}; } %
    \node[empty_block] at (6, 2) {}; %
    \foreach \i in {10,11,12} { \node[token] at (\i-1, 2) {$t_{\i}$}; } %

    \node[label] at (0, 1.3) {$q_3$};
    \foreach \i in {1,2,3} { \node[token] at (\i-1, 1) {$t_{\i}$}; } %
    \foreach \i in {4,5,6} { \node[token] at (\i-1, 1) {$t_{\i}$}; } %
    \node[empty_block] at (6, 1) {}; %
    \node[empty_block] at (9, 1) {}; %

    \node[label] at (0, 0.3) {$q_4$};
    \node[empty_block] at (0, 0) {}; %
    \node[empty_block] at (3, 0) {}; %
    \foreach \i in {7,8,9} { \node[token] at (\i-1, 0) {$t_{\i}$}; } %
    \foreach \i in {10,11,12} { \node[token] at (\i-1, 0) {$t_{\i}$}; } %

    \draw[gray!80!black, line width=1.5pt, rounded corners=4pt] 
        (0,0) rectangle (12,4);

\end{tikzpicture}
            }
            \caption{Existing methods. Teal: dense attention; White: no-op.}
            \label{fig:nsa_token}
        \end{subfigure}

    \end{minipage}

    \caption{\textit{(a) Block-level selection} illustrates the initial identification of relevant blocks. \textit{(b)} \textit{\sysname} performs exact attention on selected blocks (teal) while processing unselected blocks via residual linear attention (purple). \textit{(c) Existing sparse attention} mechanisms (e.g., InfLLM-v2, NSA) typically either discard unselected blocks entirely or process all blocks redundantly.}
    \label{fig:mixed_layout}
\end{figure*}

\subsection{Principled Block Selection via Taylor Expansion}

To solve the ungrounded selection problem, we establish an explicit mathematical connection between token-level attention and block-level importance.
We define the block relevance score as the integrated probability mass of all tokens within a block, which serves as a faithful proxy for the block's contribution to the global output:
\begin{align}
    \blockscore(\vq, \vk_{[i]}) & \propto \sum_{j \in \text{block } i} \exp(\vq \vk_j^\intercal) \\ & = B \times \E_{j} [\exp(\vq \vk_j^\intercal)] \quad \in \R^{G}.
    \label{eq:block-score-expand}
\end{align}
where $\E_{j}$ denotes the mean of the unnormalized attention scores, and the global normalization term is defined as:
\begin{equation}
    \gZ(\vq, \vk_{1:N}) = \sum_{i=1}^{N/B} B \times \E_{j} [\exp(\vq \vk_j^\intercal)] \quad \in \R^{G} \label{eq:block-score-expand-z}.
\end{equation}
Directly computing this sum is expensive. However, we show that it can be efficiently estimated via a second-order Taylor expansion around the block mean $\bar{\vk}$:
\begin{align}
\E_{\vk}[\exp(\vq^\intercal \vk)] & \approx \E_{\vk}\big[\exp(\vq^\intercal \bar \vk) \big(1 + \vq^\intercal (\vk - \bar \vk) + \nonumber \\ & \hspace{-12mm} \frac{1}{2} \vq^\intercal (\vk - \bar \vk) (\vk - \bar \vk)^\intercal \vq\big) \big] \nonumber \\
&= \exp(\vq^\intercal \bar \vk) \left(1 +  \frac{1}{2} \vq^\intercal \cov(\vk) \vq \right) 
\label{eq:taylor-2nd}  %
\\
&\approx \exp(\vq^\intercal \bar \vk), \hspace{8mm} \text{first-order}
\label{eq:taylor-1st}
\end{align}
where $\cov(\vk)$ denotes the covariance matrix.
This derivation provides a \textit{principled} selection metric: it utilizes both the first-order (mean) and second-order (variance) statistics of a block to estimate its total mass, ensuring high recall without requiring auxiliary training.
We approximate the covariance matrix $\cov(\vk)$ as diagonal to avoid the $\mathcal{O}(d^2)$ storage overhead, allowing us to compute selection metrics using only element-wise operations on block statistics.
By this approximation, we can compute both Eq~\ref{eq:block-score-expand} and Eq~\ref{eq:block-score-expand-z} efficiently.

\paragraph{Strided Block Selection}
To balance memory continuity (large $B$) with selection precision (small granularity), we adopt the strided strategy from \citet{nsa,inf-v2}.
We apply the Taylor approximation on sliding windows of size $C=32$ with a stride of $s=16$, and aggregate scores via max-pooling to determine the relevance of blocks (typically $B=128$ on TPU, $B=64$ on GPU).

\subsection{Residual Linear Attention (RLA)}

To address the "long tail divergence," we must account for the unselected blocks ($i \notin \gI$).
Instead of truncating them, we process them using a specialized linear attention mechanism.
This choice is motivated by the need for efficiency: while the "peaks" require exact attention, the "tail" consists of diffuse probability mass that can be adequately compressed into a recurrent state.

\paragraph{Recurrence}
The RLA process accumulates information exclusively from unselected blocks:
\begin{align}
 \rmS_t &= \rmS_{t-1} + \rmI_t[ \phi(\vk_t)^\intercal \vv_t] \ \ \in \R^{d\times d}, \\ \vo^{\text{\scriptsize{rla}}}_t &=  \phi(\vq_t) \rmS_{t} \ \ \in \R^{d} .
\label{eq:rla-naive}
\end{align}
where $\rmI_t$ is an identity function if $t$ is in an unselected block, and 0 otherwise. Unlike standard linear attention where state $S_t$ is shared, here the 'unselected' set is query-dependent. A naive realization would require re-scanning memory for every query, defeating the purpose of sparsity.

\paragraph{IO-Efficient Implementation}
We observe that the unselected set is the complement of the selected set. Exploiting the linearity of the operator, we derive the residual output $\vo^{\text{\scriptsize{rla}}}$ as the difference between a global context estimate and a local correction term:
\begin{align}
\vo^{\text{\scriptsize{rla}}}_t &= \bar \vo_t - \tilde \vo_t \ .
\label{eq:rla-impl-3}
\end{align}
where $\bar{\rmS}_t$ tracks the global history (computed via standard linear attention) and $\tilde{\rmS}_t$ aggregates only the currently selected blocks. Crucially, $\tilde{\rmS}_t$ is computed within the sparse kernel using blocks already loaded to SRAM. Thus, \sysname captures the full context without ever explicitly accessing unselected blocks during the sparse pass.

Here, $\bar \vo_t$ and $\tilde \vo_t$ are computed via two separate recurrences:
\begin{align}
\bar \rmS_t &=  \bar \rmS_{t-1} +  \phi(\vk_t)^\intercal \vv_t, \quad  \bar \vo_t =  \phi(\vq_t) \bar \rmS_{t},  \label{eq:rla-impl-1} \\
\tilde \rmS_t &= \tilde  \rmS_{t-1} + \tilde \rmI_t[ \phi(\vk_t)^\intercal \vv_t], \quad  \tilde \vo_t =  \phi(\vq_t) \tilde \rmS_{t}, \label{eq:rla-impl-2} 
\end{align}
Crucially, the second recurrence ($\tilde \rmS_t$) aggregates only the \textit{selected} blocks. Since these blocks are already loaded into SRAM for the exact sparse attention computation, $\tilde \vo_t$ can be computed within the same fused kernel with negligible overhead. Consequently, \sysname captures the full global context without ever explicitly accessing unselected blocks from HBM during the sparse pass.

For numerical stability, we found that as long as hidden states $\bar \rmS$ and $\tilde \rmS$ remain in high precision (i.e., float32), the output can be stable in lower precision (e.g., bfloat16). 

\paragraph{Parameterization}
Since our primary focus is adapting pretrained dense attention models via continual pretraining, we design \sysname to be fully compatible with standard attention architectures. Following \citet{rattention}, we employ a \textit{shared parameterization} scheme where the query, key, and value projections are identical for both the sparse and residual linear attention branches. The only transformation applied is the feature map function $\phi$, which projects the shared vectors into the linear attention space. In this work, we adopt a parameter-free exponential feature map $\phi(x) \sim \exp(x)$ \cite{lolcats}, ensuring that no new projection weights are introduced. This design inherently preserves the Grouped Query Attention (GQA) structure of the base model, sharing key and value heads within each query group to maintain efficiency.

While we focus on adaptation, we note that for training from scratch, further gains may be realized by decoupling these branches. Preliminary experiments suggest potential benefits from: (1) learning specialized projections for the residual branch via a parameterized feature map, or (2) employing distinct grouping strategies such as the multi-value structure proposed in \citet{mamba2}.

\paragraph{Numerical Stability}
In practice, we utilize a normalized feature map $\phi(x) = \softmax(x)$ that adds additional normalization on top of the exponential function along the feature dimension. We also found that as long as the hidden states $\bar \rmS$ and $\tilde \rmS$ are accumulated in high precision (i.e., float32), the final output remains stable even when cast to lower precision (e.g., bfloat16).

\subsection{\sysname: Sparse Plus Linear Attention}

We synthesize the exact local attention and the approximate residual attention via a gated residual connection:
\begin{align*}
    \vo_t = \vo^{\text{\scriptsize{sparse}}}_t  + \rms ( \vo^{\text{\scriptsize{rla}}}_t ) \ .
\end{align*}
where $\vo^{\text{\scriptsize{rla}}}_t = \bar \vo_t - \tilde \vo_t$. An RMS Norm is applied to the RLA output to align its magnitude with the sparse attention heads. Crucially, the scale parameters of this norm are the \textit{only} additional parameters introduced by \sysname compared to the original dense model. As shown in Section~\ref{sec:experiments}, this minimal parameterization enables \sysname to be effectively adapted from pretrained dense attention models with high sample efficiency.

We refer to this method as \textbf{S}parse \textbf{P}lus \textbf{L}inear \textbf{A}ttention (SPLA).
This intra-head hybrid strategy is in the same spirit as \citet{rattention} but differs fundamentally from other layer-wise hybrid strategy~\cite{samba,hymba,zamba}.

\paragraph{Kernel Design}
To maximize throughput, we execute two parallel kernels for both training and inference. The first is a \textit{fused sparse kernel} that handles the dynamic sparsity: it computes the exact attention $\vo^{\text{\scriptsize{sparse}}}_t$ and the selected linear term $\tilde \vo_t$ in a single pass, efficiently compressing information from selected blocks. The second is a standard \textit{linear attention kernel} that tracks the global context $\bar \vo_t$.
During inference, this decomposition offers significant efficiency gains for long contexts: the global linear kernel operates in $\mathcal{O}(1)$ step complexity by maintaining the recurrent state $\rmS_t$, while the sparse kernel processes only the top-$k$ relevant blocks, decoupling latency from the total sequence length.

\subsection{Comparison with Existing Sparse Attentions}

In terms of the general strategy for handling long contexts, \sysname distinguishes itself from existing sparse attention mechanisms through a strict partitioning approach, whereas prior works rely on either overlapping branches or aggressive truncation:
\begin{itemize}
    \item \textbf{Partitioning (\sysname):} We explicitly partition the context into disjoint sets: Exact ($i \in \gI$) and Approximate ($i \notin \gI$). By summing these components, \sysname approximates the full dense attention theoretical total without redundancy or loss.
    \item \textbf{Overlapping (NSA~\cite{nsa}):} NSA employs a multi-branch architecture where the compressed branch aggregates over \textit{all} blocks ($i \in \{1 \dots N/B\}$). This results in double-counting the information from tokens in $\gI$ (once in the exact branch, once in the compressed branch), potentially skewing the output distribution.
    \item \textbf{Truncation (InfLLM-v2~\cite{inf-v2}):} InfLLM-v2 effectively treats the unselected set as zero ($\vo_{i \notin \gI} = 0$). This is equivalent to restricting \sysname to only its sparse branch, a strategy that suffers from severe probability mass loss as context length increases.
\end{itemize}
As demonstrated empirically in Section~\ref{sec:experiments}, the partitioning strategy employed by \sysname proves significantly more robust than overlapping or truncation strategies, particularly when scaling to long contexts (up to 256k tokens).
\section{Experiments}
\label{sec:experiments}

We evaluate our sparse attention framework within a Continual Pretraining (CPT) setting. Specifically, we first pretrain a 14B dense attention model on 10 trillion tokens and subsequently adapt it into a sparse architecture during the CPT phase.
\begin{table}[t]  %
\centering
\caption{Model specifications of our 14B model.}
\begin{tabular}{l|cc}
\hline
\textbf{Parameters} &  \textbf{14B}  \\
\hline
d\_model & 5120 \\
layers &  40 \\
num heads &  40 \\
num kv heads & 8  \\
qk-norm &  yes \\
head type & GQA  \\
head size & 128 \\
rope theta & 5e5 \\
non-linearity &  GeGLU  \\
feedforward dim & 17408 \\
pre-norm  & yes \\
\hline
\end{tabular}
\vspace{-5mm}
\label{tab:model_spec}
\end{table}

Following standard protocols, our CPT stage incorporates high-quality domain-specific data (e.g., code, mathematics) and utilizes extended sequence lengths to enhance long-context capabilities. Crucially, we integrate the sparsification process directly into this stage, obviating the need for a separate post-training adaptation phase.

\paragraph{Pretraining Setup}
We conduct our experiments at the 14B parameter scale; detailed model specifications\footnote{We selected this scale to balance resource manageability with the capability for large-scale experimentation. Additionally, we aligned the specifications closely with Qwen3 14B~\cite{qwen3} to ensure our base model is not biased toward subsequent sparsification.} are provided in Table~\ref{tab:model_spec}. 
Training utilizes a global batch size of 32 million tokens with a context length of 8,192. All models are implemented in JAX~\cite{jax2018github} and trained on Cloud TPU v6e clusters. Specifically, we utilize 1024 chips arranged as $8 \times 128$-chip slices. For distributed training, we employ a combination of Fully Sharded Data Parallel (FSDP) and Sequence Parallelism (SP). The models are optimized using AdamW. We pretrain on a dataset of 10 trillion tokens, derived from internal web-crawled data with a mixture distribution following Llama~\cite{touvron2023llama}.

\paragraph{Continual Pretraining}
We establish three CPT settings to evaluate the diverse capabilities of sparse attention models:
1) \textit{General Knowledge}: Utilizing a mixture of math, code, and general knowledge domains with a context length of 32k;
2) \textit{Long-Context}: Comprising long-form documents and synthetic question-answering pairs with a context length of 128k; and
3) \textit{Reasoning}: Utilizing open-source reasoning datasets targeting complex reasoning tasks with a context length of 32k.

\paragraph{Sparsity Specifications}
For all models, we set the block size to $B=128$. This choice is driven by the TPU hardware architecture, which utilizes an $8 \times 128$ tiling layout for optimal vector processing—a deviation from prior GPU-centric works that typically default to $B=64$.\footnote{On GPU, we believe \sysname can potentially be more effective due to the capability for more fine-grained block selection.}
Regarding selection hyperparameters, we employ a compression window of $C=32$ with a stride of $s=16$. We fix the retrieval budget to the top-$32$ blocks, which we identified as an empirical sweet spot. Furthermore, following \citet{inf-v2}, we mandate the selection of the initial block (attention sink) and the most recent $4$ blocks (local sliding window), as empirical studies~\cite{minfer} indicate these positions consistently retain significant probability mass.

\paragraph{Baselines}
We compare \sysname against a full-attention baseline and two representative sparse attention baselines, \textbf{NSA} and \textbf{InfLLM-v2}, which are closely related to our method. To ensure a rigorous comparison, we fix all sparsity hyperparameters across models and implement all baselines in JAX on TPUs using a unified codebase. 
For block selection, both NSA and InfLLM-v2 use block means (rather than a learned compression network) to compute block representations. Note that in our implementation, InfLLM-v2 is effectively \sysname without the residual linear attention module, using only a first-order Taylor expansion for block selection.

To further verify our design choices, we include an ablation variant denoted as \textbf{SPA}. SPA utilizes the same second-order Taylor-based block selection as \sysname but excludes the Residual Linear Attention module. This setup allows us to disentangle the sources of improvement: comparing \sysname with SPA isolates the specific benefits of incorporating residual linear attention, while comparing SPA with InfLLM-v2 highlights the efficacy of our principled block selection mechanism.

\subsection{Evaluation Results}

We compare \sysname with the baselines across the three CPT settings.

\begin{table}[t]
\centering
\caption{Results on general knowledge benchmarks.}
\resizebox{0.5 \textwidth}{!}{%
\begin{tabular}{l|c||c|c||c|c}
\toprule
\textbf{Metric} & \textsc{Dense} & \textsc{Inf-v2}   & \textsc{NSA} & \sysnameablate &  \sysname  \\
\hline
ARC-C & 60.1 & 59.2 & 57.0 & 58.9 & 60.2  \\
ARC-E & 85.3 & 85.5 & 85.5 & 85.2 & 85.5  \\
HellaSwag & 61.8 & 61.6 & 60.1 & 61.8 &  61.9  \\
LAMBADA & 74.9 & 75.3 & 73.5 & 75.1 & 75.2 \\
PIQA & 80.9 & 80.5 & 79.8 & 80.3 & 80.6  \\
SciQ & 95.8 & 96.2 & 95.6 & 96.2 & 95.8  \\
WinoGrande & 74.0 & 74.4 & 73.8 & 74.5  & 75.0 \\
TriviaQA (1-shot) & 49.6 & 49.1 & 49.3 & 49.8  & 49.6 \\
WebQS (1-shot) & 24.9 & 23.9 & 24.7 & 24.9 & 25.6 \\
\rowcolor{gray!20} \textbf{Average (0/1-shot)} & 67.5 & 67.3 & 66.5 & 67.3  & \textbf{67.7}  \\
\hline
\rowcolor{gray!20} MMLU (5-shot) & 77.3 & 77.8 & 77.4 & 78.1  & \textbf{78.6} \\
\rowcolor{gray!20} GSM8K (8-shot) & 90.2 & 90.0 & 90.5 & 90.2  &  90.5  \\
\toprule
\end{tabular}%
}
\label{tab:general_results}
\end{table}
\begin{table}[t!]
\centering
\caption{Long-context performance on RULER benchmarks.}
\resizebox{0.5 \textwidth}{!}{%
\begin{tabular}{l|c|c|c|c|c|c|c}
\hline
\textbf{Model} & \textbf{4k} & \textbf{8k} & \textbf{16k} & \textbf{32k} & \textbf{64k} & \textbf{128k} & \textbf{256k} \\
\hline
\textsc{Dense} & 95.8 & 94.9 & 93.6 & 91.4 & 87.1 & 83.2 & 69.3 \\
\textsc{NSA} & 92.3 & 91.4 & 90.7 & 84.6 & 83.2 & 51.3 & 32.5  \\
\textsc{Inf-v2} &  
95.9 & 94.1 & 92.1 & 89.3 & 86.7 & 61.6 & 42.6  \\
\hline
\sysnameablate & 95.9 & 94.2  & 93.3 & 90.4 & 87.1 & 62.4 & 44.5  \\
\sysname & 95.9 & 94.7  & 94.2 & 91.7 & \textbf{88.3} & \textbf{85.2} & \textbf{72.3} \\
\toprule
\end{tabular}%
}
\vspace{-3mm}
\label{tab:ruler_results}
\end{table}

\paragraph{General Knowledge}
In this setting, we continually pretrain models on our in-house high-quality data featuring math, code, and knowledge at a context length of 32k for 640B tokens. For evaluation, we consider a standard suite of tasks: SciQ~\cite{welbl2017sciq}, TriviaQA~\cite{joshi2017triviaqa}, WebQ~\cite{berant-etal-2013-webqs}, MMLU~\cite{hendrycks2020mmlu}, GSM8k~\cite{cobbe2021gsm8k}, LAMBADA~\cite{paperno2016lambada}, PiQA~\cite{bisk2020piqa}, HellaSwag~\cite{zellers2019hellaswag}, WinoGrande~\cite{sakaguchi2021winogrande}, ARC-Easy (ARC-E), and ARC-Challenge (ARC-C)~\citep{arc-ce}. The results are shown in Table~\ref{tab:general_results}.

\paragraph{Long-Context}
In this setting, we use our curated long-context dataset containing long documents along with synthetic question-answering pairs. All models are trained with a context length of 128k for 130B tokens. We evaluate both in-distribution generalization ($ \leq 128k$) and extrapolation ($256k$) capability on the RULER~\cite{hsieh2024ruler} benchmark. Results are shown in Table~\ref{tab:ruler_results}.

\paragraph{Reasoning}
In this setting, we follow the OpenReasoning Nemotron recipe\footnote{\url{https://huggingface.co/nvidia/OpenReasoning-Nemotron-14B}} using a mixture of math and code data from \citet{NemotronPostTrainingDatasetV1} and science data from OpenScienceReasoning\footnote{\url{https://huggingface.co/datasets/nvidia/OpenScienceReasoning-2}}. All models are trained for 1.2T tokens with a context length of 32k. We evaluate models on American Invitational Mathematics Examinations (AIME), Harvard-MIT Mathematics Tournaments (HMMT), LiveCodeBench~\cite{livecodebench} (release-v5), HumanEval~\cite{humaneval}, MMLU Pro~\cite{mmlupro}, and GPQA (Diamond)~\cite{gpqa}. Decoding employs a temperature of 0.6, a maximum output length of 32k tokens, and a single sample (pass$@1$). The results are reported in Table~\ref{tab:reasoning_results}.

\begin{table}[t!]
\centering
\caption{Results on reasoning and knowledge benchmarks. }
\resizebox{0.5 \textwidth}{!}{%
\begin{tabular}{l|c||c|c||c|c}
\textbf{Metric} & \textsc{Dense} & \textsc{\ NSA\ } & \textsc{Inf-v2} & \sysnameablate & \sysname  \\
\hline
AIME 2024 & 77.1 & 73.1 & 76.8 & 77.2 & \textbf{78.3}  \\
AIME 2025 & 73.9 &  64.2 & 75.1 &  73.0 & 75.1  \\
HMMT 2025 & 60.0 & 53.3 & 60.0 & 60.0& \textbf{63.3} \\
LiveCodeBench  & 61.6 & 49.1  & \textbf{64.2} &  62.0 & 62.4  \\
HumanEval & 85.4 &  78.7  & 86.6 & 86.0 & 86.6  \\
GPQA & 68.5 & 59.6 & 68.7 & 69.2 & \textbf{69.5} \\
MMLU Pro & 78.9 & 68.8 & 79.3 & 79.1 & 79.3   \\

\hline
\end{tabular}%
}
\vspace{-2mm}
\label{tab:reasoning_results}
\end{table}

\begin{figure*}[t]
\centering
\includegraphics[width=0.9 \textwidth]{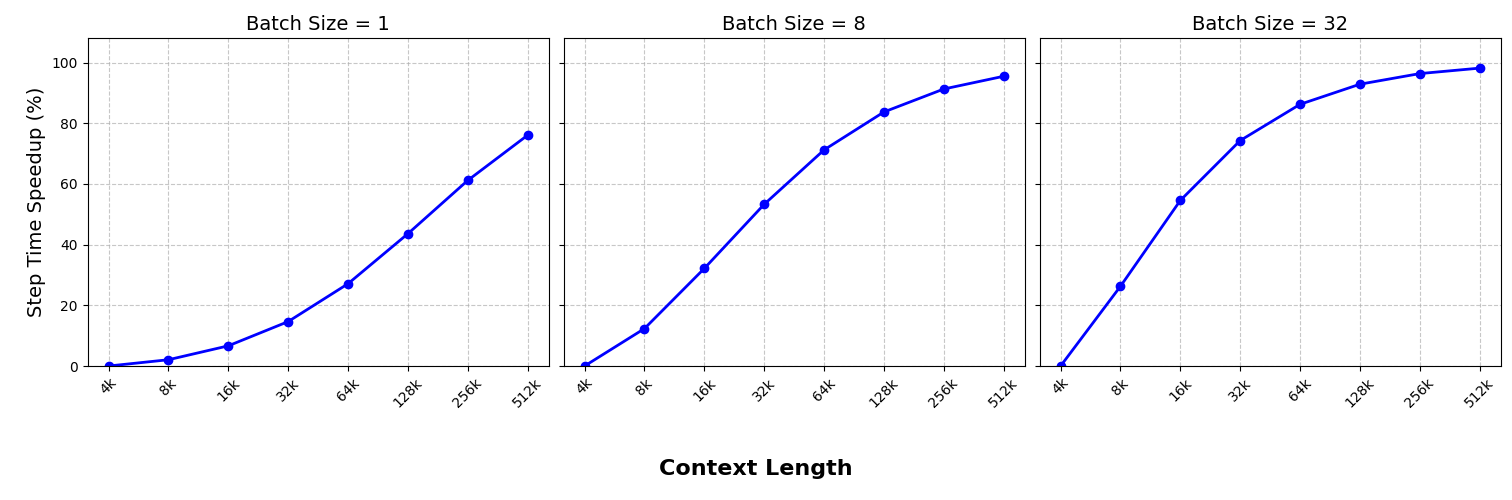}
\vspace{-2mm}
\caption{Decoding step time speedup (\%) of \sysname models compared to dense-attention models at 13B scale. As the batch size increases, the theoretical speedup of \sysname increases more quickly, since the KV cache size increasingly dominates the memory cost relative to the model parameter size.}
\label{fig:step_time}
\vspace{-3mm}
\end{figure*}

\subsection{Discussion}

\paragraph{Impact of Residual Linear Attention}
Across all three benchmarks, \sysname consistently outperforms the ablated variant SPA, demonstrating the critical effectiveness of the Residual Linear Attention (RLA) module. By compressing the ``long tail'' rather than discarding it, \sysname prevents the divergence that typically plagues sparse models at scale. This is most evident on the long-context benchmarks, where \sysname proves particularly robust, achieving even stronger performance than dense attention baselines as sequence lengths exceed 64k.

\paragraph{Selection Fidelity and Reasoning}
Regarding block selection, the ablation variant SPA generally surpasses InfLLM-v2, validating the superiority of our principled Taylor-based strategy over standard first-order approximations. We also observe that NSA exhibits significantly weaker performance on the reasoning benchmark. We attribute this to the use of a simplified block compressor (mean pooling) in the baseline implementation, which likely fails to retain the fine-grained information necessary for complex reasoning tasks—a limitation that \sysname overcomes through its hybrid exact-plus-linear design.

\subsection{Analysis}

\paragraph{Training Efficiency}
For large-scale experimentation, we developed a general sparse attention kernel on TPU compatible with \sysname{}, NSA, and InfLLM-v2. Leveraging this unified implementation, all sparse variants exhibit comparable training throughput. Compared to the highly optimized dense attention baseline, we currently observe a computational overhead. Specifically, at a 16k context length, the step time is 11s for sparse models versus 9s for dense models. We attribute this latency primarily to the overhead of the block selection mechanism and the relatively small GQA group size ($G=5$) used in our 14B model configuration, which limits parallelization efficiency. With a larger group size ($G=8$ or $16$), the training speed is roughly on par. Overall, \sysname follows the general trend of sparse attention models preferring larger group sizes. Crucially, this training efficiency demonstrates that \sysname is viable for training from scratch at large scale.

\paragraph{Inference Efficiency}
We analyze the inference latency during the decoding phase, focusing on the scalability of sparse mechanisms relative to dense attention. It is worth noting that \sysname share a similar computational profile with other block-sparse baselines (e.g., NSA~\cite{nsa}, InfLLM-v2~\cite{inf-v2}): the primary efficiency gain stems from reducing the memory I/O required to load the KV cache.
During decoding, the attention operation is heavily memory-bound. While the arithmetic intensity of dense attention scales linearly with sequence length, the data transfer cost dominates total latency.
As illustrated in Figure~\ref{fig:step_time}, sparse attention mechanisms break this bottleneck by capping the memory access to a fixed number of selected blocks (plus the negligible overhead of the recurrent state updates in \sysname).
Consequently, while gains are modest at short contexts (where loading model weights dominates), the speedup becomes substantial as the context length and batch size increase, causing the KV cache footprint to become the primary bottleneck.
\section{Related Work}

We categorize existing work into three types based on whether training is needed and whether a separate proxy attention is employed.

\paragraph{Training Free Sparse Attention}
Early sparse attention works are only employed as inference techniques without changing the training or architecture of models. Although limited in generalization, these works shed light on intrinsic patterns of dense attentions. For example, StreamingLLM~\cite{streaming} proposed the static pattern of attending to early "sink" tokens as well as a recent sliding window of tokens. Although such patterns fail to generalize~\cite{inf-v1}, they still form the basic inductive biases for later sparse attention models, including \sysname. MInference~\cite{minfer} found a diverse set of patterns in dense attention which are crucial for approximation. Quest~\cite{quest} emphasizes the importance of sparsity patterns being highly input-dependent.

\paragraph{Learning-based Sparse Attention} 
On the other end of the spectrum are learning-based methods where proxy efficient attention mechanisms are used for selection, instead of directly relying on the original dense attention. DSA~\cite{dsa} employs a cheaper attention, namely a lightning indexer, to choose relevant tokens for the dense attention to perform computation. SeerAttention~\cite{seer-attn} proposed to learn a blockwise attention for block selection. The learning signals are based on the distance (often KL divergence) between the proxy attention and the target dense attention. In general, we believe that learning-based methods enjoy better flexibility with extra parameters for sparsity learning. However, the learning part poses significant practical challenges, e.g., how to effectively learning those parameters in continued pretraining settings along with other parameters.

\paragraph{Approximate Sparse Attention}Distinct from learned proxies, approximation-based methods aim to derive a sparse representation directly from the original dense attention statistics. This category includes TopK selection methods such as NSA~\cite{nsa}, MoBA~\cite{moba} and Inf-LLM v2~\cite{inf-v2}. As detailed in Section~\ref{sec:background}, these methods suffer from a "hard truncation" error: by retaining only the top-$k$ blocks and zeroing out the rest, they discard the heavy tail of the distribution, which becomes significant in long-context regimes. Attempts to mitigate this include TopP selection~\cite{tactic}, which ensures better mass coverage but incurs unpredictable computational costs. Alternatively, sampling-based methods like MagicPig~\cite{magicpig}, SampleAttention~\cite{sampleattention}, and vAttention~\cite{vattention} employ sampling to estimate the contribution of unselected regions. However, sampling would presumably suffer from high variance as context length grows. \sysname represents an advancement within this category. Rather than relying on hard truncation or stochastic sampling, we employ a parameter-efficient residual linear attention to compress all unselected blocks. 

\section{Conclusion}
In this work, we propose \sysname, a novel sparse attention mechanism that advances sparse attention techniques in two critical aspects:
principled block selection mechanism and the efficient compression of unselected blocks using residual linear attention.
Through comprehensive empirical evaluation, we demonstrate that pretrained dense attention models can be successfully adapted to \sysname architectures during the continual pretraining stage.
Crucially, this adaptation preserves or even improves performance across general knowledge, reasoning, and long-context benchmarks,
offering a scalable pathway for serving next-generation foundation models.

\section*{Acknowledgment}

We thank Yihao Feng, Sam Wiseman and Jianyu Wang for their valuable feedback.

\bibliography{main}
\bibliographystyle{icml2026}

\end{document}